\documentclass[10pt,twocolumn,letterpaper]{article}

\usepackage{iccv}
\usepackage{times}
\usepackage{epsfig}
\usepackage{graphicx}
\usepackage{amsmath}
\usepackage{amssymb}

\usepackage{caption}
\usepackage{subcaption}
\usepackage{booktabs}
\usepackage{array,multirow}
\usepackage{xcolor}
\usepackage[british,american]{babel}
\usepackage{blindtext}
\usepackage[inline]{enumitem}
\usepackage{balance}
\usepackage{soul}
\usepackage{authblk}
\usepackage[pagebackref=true,breaklinks=true,letterpaper=true,colorlinks,bookmarks=false]{hyperref}
\usepackage{textpos}

\iccvfinalcopy

\ificcvfinal\fi

\newcommand{\figref}[1]{\mbox{Fig.~\ref{#1}}}
\newcommand{\tblref}[1]{\mbox{Table~\ref{#1}}}
\newcommand{\secref}[1]{\mbox{Sec.~\ref{#1}}}
\renewcommand{\eqref}[1]{\mbox{Eq.~\ref{#1}}}

\newcommand{\ours}{\mbox{{Ours}~}}

\newif\ifedit

\usepackage[font=small]{subcaption}
\let\oldparagraph\paragraph
\renewcommand{\paragraph}[1]{\vspace{-0.4cm} \oldparagraph{#1}}
\newcommand{\abstrctvspace}{\vspace{-0.3cm}}
\newcommand{\figvspace}{\vspace{-0.4cm}}
\newcommand{\figlblvspace}{\vspace{-0.3cm}}
\newcommand{\eqtopvspace}{\vspace{-0.1cm}}
\newcommand{\eqbottomvspace}{\vspace{-0.1cm}}
\newcommand{\tabvspace}{\vspace{-0.2cm}}
\newcommand{\tablblvspace}{\vspace{-0.2cm}}

\begin{document}

\title{Fashion Forward: Forecasting Visual Style in Fashion}

\newcommand*{\affmark}[1][*]{\textsuperscript{#1}}
\newcommand*{\email}[1]{\tt\small{#1}}
\author{
Ziad Al-Halah\affmark[1] \hspace{1cm} Rainer Stiefelhagen\affmark[1] \hspace{1cm} Kristen Grauman\affmark[2]\\
\affmark[1]Karlsruhe Institute of Technology, 76131 Karlsruhe, Germany\\
\affmark[2]The University of Texas at Austin, 78701 Austin, USA\\
\email{ziadlhlh@gmail.com \hspace{.2cm} rainer.stiefelhage@kit.edu \hspace{.2cm} grauman@cs.utexas.edu}
}

\maketitle

\begin{abstract}

What is the future of fashion?
Tackling this question from a data-driven vision perspective, we propose to forecast visual style trends before they occur.
We introduce the first approach to predict the future popularity of styles discovered from fashion images in an unsupervised manner.
Using these styles as a basis, we train a forecasting model to represent their trends over time.
The resulting model can hypothesize new mixtures of styles that will become popular in the future, discover style dynamics (trendy vs.~classic), and name the key visual attributes that will dominate tomorrow's fashion.
We demonstrate our idea applied to three datasets encapsulating 80,000 fashion products sold across six years on Amazon.
Results indicate that fashion forecasting benefits greatly from \emph{visual} analysis, much more than textual or meta-data cues surrounding products. 
Project page: \url{https://cvhci.anthropomatik.kit.edu/~zalhalah/prj_fashion_forecast.html}.
 \end{abstract}
\abstrctvspace
\begin{textblock*}{\textwidth}(0cm,-15cm)
\centering\small
IEEE International Conference on Computer Vision (ICCV), 2017.
\end{textblock*}
\section{Introduction}

\emph{``The customer is the final filter. What survives the whole process is what people wear.''} -- Marc Jacobs
\vspace*{0.05in}

Fashion is a fascinating domain for computer vision.
Not only does it offer a challenging testbed for fundamental vision problems---human body parsing~\cite{paperdoll-iccv2013,yamaguchi-cvpr2012}, cross-domain image matching~\cite{Liu2012Street,Kiapour2015,Huang2015,Chen2015DeepDomain}, and recognition~\cite{Bossard2012,Liu2016,Chen2012,Kiapour2014}---but it also inspires new problems that can drive a research agenda, such as modeling visual compatibility~\cite{Iwata2011,Veit2015}, interactive fine-grained retrieval~\cite{Kovashka2012,Yu2015}, or reading social cues from what people choose to wear~\cite{Kwak2013,Song2011,Chen2015Devils,Simo-Serra2015}.
At the same time, the space has potential for high impact: the global market for apparel is estimated at \$3 Trillion USD~\cite{fashionunited}.
It is increasingly entwined with online shopping, social media, and mobile computing---all arenas where automated visual analysis should be synergetic.

\begin{figure}[t]
\centering
    \includegraphics[width=\linewidth]{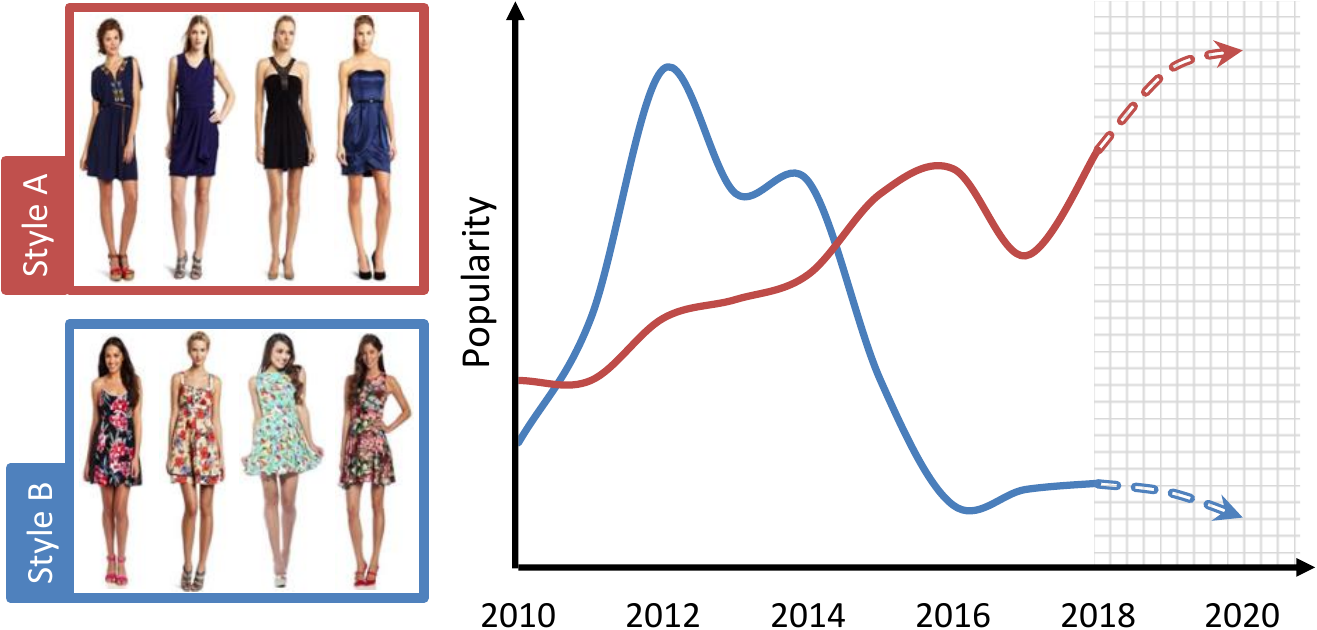}
\figlblvspace
\caption{We propose to predict the future of fashion based on visual styles.}
\label{fig:intro}
\figvspace
\end{figure}

In this work, we consider the problem of \emph{visual fashion forecasting}.
The goal is to predict the future popularity of fine-grained fashion styles.
For example, having observed the purchase statistics for all women's dresses sold on Amazon over the last $N$ years, can we predict what salient visual properties the best selling dresses will have 12 months from now?
Given a list of trending garments, can we predict which will remain stylish into the future?
Which old trends are primed to resurface, independent of seasonality?

Computational models able to make such forecasts would be critically valuable to the fashion industry, in terms of portraying large-scale trends of what people will be buying months or years from now.
They would also benefit individuals who strive to stay ahead of the curve in their public persona, e.g., stylists to the stars.
However, fashion forecasting is interesting even to those of us unexcited by haute couture, money, and glamour.
This is because wrapped up in everyday fashion trends are the effects of shifting cultural attitudes, economic factors, social sharing, and even the political climate.
For example, the hard-edged flapper style during the prosperous 1920's in the U.S. gave way to the conservative, softer shapes of 1930's women's wear, paralleling current events such as women's right to vote (secured in 1920) and the stock market crash 9 years later that prompted more conservative attitudes~\cite{DavisBook}.
Thus, beyond the fashion world itself, quantitative models of style evolution would be valuable in the social sciences.

While structured data from vendors (i.e., recording purchase rates for clothing items accompanied by meta-data labels) is relevant to fashion forecasting, we hypothesize that it is not enough.
Fashion is visual, and comprehensive fashion forecasting demands actually looking at the products.
Thus, a key technical challenge in forecasting fashion is how to represent visual style.
Unlike articles of clothing and their attributes (e.g., sweater, vest, striped), which are well-defined categories handled readily by today's sophisticated visual recognition pipelines~\cite{Bossard2012,Chen2012,Liu2016,Simo-Serra2016}, styles are more difficult to pin down and even subjective in their definition.
In particular, two garments that superficially are visually different may nonetheless share a style.

Furthermore, as we define the problem, fashion forecasting goes beyond simply predicting the future purchase rate of an individual item seen in the past.
So, it is not simply a regression problem from images to dates.
Rather, the forecaster must be able to hypothesize
styles that will \emph{become} popular in the future---\ie, to generate yet-unseen compositions of styles.
The ability to predict the future of styles rather than merely items is appealing for applications that demand interpretable models expressing where trends as a whole are headed, as well as those that need to capture the life cycle of collective styles, not individual garments. 
Despite some recent steps to qualitatively analyze past fashion trends in hindsight~\cite{Vittayakorn2015,Simo-Serra2015,Chen2015Devils,Vittayakorn2016,He2016}, to our knowledge no existing work attempts visual fashion forecasting.

We introduce an approach that forecasts the popularity of visual styles discovered in unlabeled images.  
Given a large collection of unlabeled fashion images, we first predict clothing attributes using a supervised deep convolutional model.
Then, we discover a ``vocabulary" of latent styles using non-negative matrix factorization.
The discovered styles account for the attribute combinations observed in the individual garments or outfits.
They have a mid-level granularity: they are more general than individual attributes (pastel, black boots), but more specific than typical style classes defined in the literature (preppy, Goth, etc.)~\cite{Kiapour2014,Veit2015,Simo-Serra2016}.
We further show how to augment the visual elements with text data, when available, to discover fashion styles.
We then train a forecasting model to represent trends in the latent styles over time and to predict their popularity in the future.
Building on this, we show how to extract style dynamics (trendy vs.~classic vs.~outdated), and forecast the key visual attributes that will play a role in tomorrow's fashion---all based on learned \emph{visual} models.

We apply our method to three datasets covering six years of fashion sales data from Amazon for about 80,000 unique products.
We validate the forecasted styles against a held-out future year of purchase data.
Our experiments analyze the tradeoffs of various forecasting models and representations, the latter of which reveals the advantage of unsupervised style discovery based on visual semantic attributes compared to off-the-shelf CNN representations, including those fine-tuned for garment classification.    
Overall, an important finding is that visual content is crucial for securing the most reliable fashion forecast.
Purchase meta-data, tags, etc., are useful, but can be insufficient when taken alone.

\section{Related work}
\vspace{0.3cm}

\paragraph{Retrieval and recommendation}

There is strong practical interest in matching clothing seen on the street to an online catalog, prompting methods to overcome the street-to-shop domain shift~\cite{Liu2012Street,Kiapour2015,Huang2015}.
Beyond exact matching, recommendation systems require learning when items ``go well" together~\cite{Iwata2011,Veit2015,Simo-Serra2015} and capturing personal taste~\cite{Bracher2016} and occasion relevance~\cite{Liu2012Magic}.
Our task is very different.
Rather than recognize or recommend garments, our goal is to forecast the future popularity of styles based on visual trends.

\paragraph{Attributes in fashion}

Descriptive visual attributes are naturally amenable to fashion tasks, since garments are often described by their materials, fit, and patterns (\emph{denim}, \emph{polka-dotted}, \emph{tight}).  
Attributes are used to recognize articles of clothing~\cite{Bossard2012,Liu2016}, retrieve products~\cite{Huang2015,Di2013}, and describe clothing~\cite{Chen2012,Chen2015DeepDomain}.
Relative attributes~\cite{Parikh2011} are explored for interactive image search with applications to shoe shopping~\cite{Kovashka2012,Yu2015}.
While often an attribute vocabulary is defined manually, useful clothing attributes are discoverable from noisy meta-data on shopping websites~\cite{Berg2010} or neural activations in a deep network~\cite{Vittayakorn2016a}.
Unlike prior work, we use inferred visual attributes as a conduit to discover fine-grained fashion styles from unlabeled images.

\paragraph{Learning styles}

Limited work explores representations of visual \emph{style}.
Different from recognizing an article of clothing (\emph{sweater}, \emph{dress}) or its attributes (\emph{blue}, \emph{floral}), styles entail the higher-level concept of how clothing comes together to signal a trend.
Early methods explore supervised learning to classify people into style categories, e.g., biker, preppy, Goth~\cite{Kiapour2014,Veit2015}.
Since identity is linked to how a person chooses to dress, clothing can be predictive of occupation~\cite{Song2011} or one's social ``urban tribe"~\cite{Kwak2013,Murillo2012}.
Other work uses weak supervision from meta-data or co-purchase data to learn a latent space imbued with style cues~\cite{Simo-Serra2016,Veit2015}.
In contrast to prior work, we pursue an unsupervised approach for discovering visual styles from data, which has the advantages of i) facilitating large-scale style analysis, ii) avoiding manual definition of style categories, iii) allowing the representation of finer-grained styles 
, and iv) allowing a single outfit to exhibit multiple styles.
Unlike concurrent work \cite{Hsiao2017} that learns styles of outfits, we discover styles for individual garments and, more importantly, predict their popularity in the future.

\paragraph{Discovering trends}

Beyond categorizing styles, a few initial studies analyze fashion \emph{trends}.  
A preliminary experiment plots frequency of attributes (floral, pastel, neon) observed over time~\cite{Vittayakorn2015}.  
Similarly, a visualization shows the frequency of garment meta-data over time in two cities~\cite{Simo-Serra2015}.
The system in~\cite{Vittayakorn2016} predicts when an object was made.
The collaborative filtering recommendation system of~\cite{He2016} is enhanced by accounting for the temporal dynamics of fashion, with qualitative evidence it can capture popularity changes of items in the past (i.e., Hawaiian shirts gained popularity after 2009).
A study in~\cite{Chen2015Devils} looks for correlation between attributes popular in New York fashion shows versus what is seen later on the street.
Whereas all of the above center around analyzing \emph{past} (observed) trend data, we propose to forecast the \emph{future} (unobserved) styles that will emerge.  
To our knowledge, our work is the first to tackle the problem of visual style forecasting, and we offer objective evaluation on large-scale datasets.

\paragraph{Text as side information}

Text surrounding fashion images can offer valuable side information.
Tag and garment type data can serve as weak supervision for style classifiers~\cite{Simo-Serra2016,Simo-Serra2015}.
Purely textual features (no visual cues) are used to discover the alignment between words for clothing elements and styles on the fashion social website Polyvore~\cite{Vaccaro2016}.
Similarly, extensive tags from experts can help learn a representation to predict customer-item match likelihood for recommendation~\cite{Bracher2016}.
Our method can augment its visual model with text, when available.
While \emph{adding} text improves our forecasting, we find that text alone is inadequate; the visual content is essential.

\section{Learning and forecasting fashion style}
We propose an approach to predict the future of fashion styles based on images and consumers' purchase data.
Our approach 1) learns a representation of fashion images that captures the garments' visual attributes;
then 2) discovers a set of fine-grained styles that are shared across images in an unsupervised manner;
finally, 3) based on statistics of past consumer purchases, constructs the styles' temporal trajectories and predicts their future trends.

 \subsection{Elements of fashion}\label{sec:app_attributes}
In some fashion-related tasks, one might rely solely on meta information provided by product vendors, \eg, to analyze customer preferences.
Meta data such as tags and textual descriptions are often easy to obtain and interpret.
However, they are usually noisy and incomplete.
For example, some vendors may provide inaccurate tags or descriptions in order to improve the retrieval rank of their products,
and even extensive textual descriptions fall short of communicating all visual aspects of a product.

On the other hand, images are a key factor in a product's representation.
It is unlikely that a customer will buy a garment without an image no matter how expressive the textual description is.
Nonetheless, low level visual features are hard to interpret.
Usually, the individual dimensions are not correlated with a semantic property.
This limits the ability to analyze and reason about the final outcome and its relation to observable elements in the image.
Moreover, these features often reside in a certain level of granularity.
This renders them ill-suited to capture the fashion elements which usually span the granularity space from the most fine and local (\eg collar) to the coarse and global (\eg cozy).

Semantic attributes serve as an elegant representation that is both interpretable and detectable in images.
Additionally, they express visual properties at various levels of granularity.
Specifically, we are interested in attributes that capture the diverse visual elements of fashion, like:
\begin{itemize*}[label={}, itemjoin={;~}]
	\item \emph{Colors} (\eg blue, pink)
	\item \emph{Fabric} (\eg leather, tweed)
	\item \emph{Shape} (\eg midi, beaded)
	\item \emph{Texture} (\eg floral, stripe); etc
\end{itemize*}.
These attributes constitute a natural vocabulary to describe styles in clothing and apparel.
As discussed above, some prior work considers fashion attribute classification \cite{Liu2016,Huang2015}, though none for capturing higher-level visual styles.

To that end, we train a deep convolutional model for attribute prediction using the DeepFashion dataset \cite{Liu2016}.
The dataset contains more than 200,000 images labeled with 1,000 semantic attributes collected from online fashion websites.
Our deep attribute model has an AlexNet-like structure \cite{Krizhevsky2012}.
It consists of 5 convolutional layers and three fully connected layers.
The last attribute prediction layer is followed by a sigmoid activation function.
We use the cross entropy loss to train the network for binary attribute prediction.
The network is trained using Adam \cite{Kingma2015} for stochastic optimization with an initial learning rate of 0.001 and a weight decay of 5e-4. (see Supp.~for details).

With this model we can predict the presence of $M=1,000$ attributes in new images:
\begin{equation}
\eqtopvspace
\mathbf{a}_i = f_a(x_i|\theta),
\eqbottomvspace
\end{equation}
such that $\theta$ is the model parameters, and $\mathbf{a}_i \in \mathbb{R}^M$ where the $m^{th}$ element in $\mathbf{a}_i$ is the probability of attribute $a^m$ in image $x_i$, \ie, $a_i^m=p(a^m|x_i)$.
$f_a(\cdot)$ provides us with a detailed visual description of a garment that, as results will show, goes beyond meta-data typically available from a vendor.
 \subsection{Fashion style discovery}\label{sec:app_style}

For each genre of garments (\eg, Dresses or T-Shirts), we aim to discover the set of fine-grained styles that emerge.
That is, given a set of images $X=\{x_i\}_{i=1}^N$ we want to discover the set of $K$ latent styles $S=\{s_k\}_{k=1}^K$ that are distributed across the items in various combinations.

We pose our style discovery problem in a nonnegative matrix factorization (NMF) framework that maintains the interpretability of the discovered styles and scales efficiently to large datasets.
First we infer the visual attributes present in each image using the classification network described above.
This yields an $M\times N$ matrix $\mathbf{A}\in \mathbb{R}^{M\times N}$ indicating the probability that each of the $N$ images contains each of the $M$ visual attributes.
Given $\mathbf{A}$, we infer the matrices $\mathbf{W}$ and $\mathbf{H}$ with nonnegative entries such that:
\begin{equation}\label{eq:nmf}
\eqtopvspace
\mathbf{A} \approx \mathbf{W} \mathbf{H}~~~~\mathrm{where}~~\mathbf{W}\in \mathbb{R}^{M\times K},~\mathbf{H}\in \mathbb{R}^{K\times N}.
\eqbottomvspace
\end{equation}
We consider a low rank factorization of $\mathbf{A}$, such that $\mathbf{A}$ is estimated by a weighted sum of $K$ rank-1 matrices:
\begin{equation}\label{eq:cp}
\eqtopvspace
\begin{split}
\mathbf{A} \approx \sum_{k=1}^K \lambda_k.\mathbf{w}_k\otimes\mathbf{h}_k,
\end{split}
\eqbottomvspace
\end{equation}
where $\otimes$ is the outer product of the two vectors and $\lambda_k$ is the weight of the $k^{th}$ factor \cite{kolda2009tensor}.

By placing a Dirichlet prior on $\mathbf{w}_k$ and $\mathbf{h}_k$, we insure the nonnegativity of the factorization.
Moreover, since $||\mathbf{w}_k||_1=1$, the result can be viewed as a topic model with the styles learned by \eqref{eq:nmf} as topics over the attributes.
That is, the vectors $\mathbf{w}_k$ denote common combinations of selected attributes that emerge as the latent style ``topics", such that $w^m_k=p(a_m|s_k)$.
Each image is a mixture of those styles, and the combination weights in $\mathbf{h}_k$, when $\mathbf{H}$ is column-wise normalized, reflect the strength of each style for that garment, \ie, $h^i_k=p(s_k|x_i)$.

Note that our style model is unsupervised which makes it suitable for style discovery from large scale data.  Furthermore, we employ an efficient estimation for \eqref{eq:cp} for large scale data using an online MCMC based approach \cite{Hu2015NTF}.
At the same time, by representing each latent style $s_k$ as a mixture of attributes $[a_k^1, a_k^2,\dots,a_k^M]$, we have the ability to provide a semantic linguistic description of the discovered styles in addition to image examples.
Figure~\ref{fig:styles_topimgs} shows examples of styles discovered for two datasets (genres of products) studied in our experiments.

Finally, our model can easily integrate multiple representations of fashion when it is available by adjusting the matrix $\mathbf{A}$.
That is, given an additional view (\eg, based on textual description) of the images $\mathbf{U}\in\mathbb{R}^{L\times N}$, we augment the attributes with the new modality to construct the new data representation $\acute{\mathbf{A}}=[\mathbf{A};\mathbf{U}] \in \mathbb{R}^{(M+L) \times N}$.
Then $\acute{\mathbf{A}}$ is factorized as in \eqref{eq:nmf} to discover the latent styles.
 \subsection{Forecasting visual style}\label{sec:app_forecast}

We focus on forecasting the future of fashion over a 1-2 year time course.
In this horizon, we expect consumer purchase behavior to be the foremost indicator of fashion trends.
In longer horizons, \eg, 5-10 years, we expect more factors to play a role in shifting general tastes, from the social, political, or demographic changes to technological and scientific advances.  
Our proposed approach could potentially serve as a quantitative tool towards understanding trends in such broader contexts, but modeling those factors is currently out of the scope of our work.

\paragraph{The temporal trajectory of a style}
In order to predict the future trend of a visual style, first we need to recover the temporal dynamics which the style went through up to the present time.
We consider a set of customer transactions $Q$ (\eg, purchases) such that each transaction $q_i\in Q$ involves one fashion item with image $x_{q_i} \in X$.
Let $Q^t$ denote the subset of transactions at time $t$, \eg, within a period of one month.
Then for a style $s_k\in S$, we compute its temporal trajectory $y^k$ by measuring the relative frequency of that style at each time step:
\begin{equation}
\eqtopvspace
y_t^k = \frac{1}{|Q^t|}\sum_{q_i \in Q^t} p(s_k|x_{q_i}),
\eqbottomvspace
\end{equation}
for $t=1,\dots,T$.
Here $p(s_k|x_{q_i})$ is the probability for style $s_k$ given image $x_{q_i}$ of the item in transaction $q_i$.

\paragraph{Forecasting the future of a style}
Given the style temporal trajectory up to time $n$, we predict the popularity of the style in the next time step  in the future $\hat{y}_{n+1}$ using an exponential smoothing model \cite{brown1961fundamental}:
\begin{equation}
\eqtopvspace
\begin{split}
\hat{y}_{n+1|n} &= l_n \\
l_n &= \alpha y_n + (1-\alpha)l_{n-1} \\
\hat{y}_{n+1|n} &= \sum_{t=1}^{n}\alpha(1-\alpha)^{n-t}y_t + (1-\alpha)^{n}l_0
\end{split}
\eqbottomvspace
\end{equation}
where $\alpha \in [0,1]$ is the smoothing factor, $l_n$ is the smoothing value at time $n$, and $l_0=y_0$.
In other words, our forecast $\hat{y}_{n+1}$ is an estimated mean for the future popularity of the style given its previous temporal dynamics.

The exponential smoothing model (EXP), with its exponential weighting decay, nicely captures the intuitive notion that the most recent observed trends and popularities of styles have higher impact on the future forecast than older observations.
Furthermore, our selection of EXP combined with $K$ independent style trajectories is partly motivated by practical matters, namely the public availability of product image data accompanied by sales rates.
EXP is defined with only one parameter ($\alpha$) which can be efficiently estimated from relatively short time series.
In practice, as we will see in results, it outperforms several other standard time series forecasting algorithms, specialized neural network solutions, and a variant that models all $K$ styles jointly (see \secref{sec:eval_forecast}).
While some styles' trajectories exhibit seasonal variations (\eg T-Shirts are sold in the summer more than in the winter), such changes are insufficient with regard of the general trend of the style.  
As we show later, the EXP model outperforms models that incorporate seasonal variations or styles' correlations for our datasets.

\section{Evaluation}

Our experiments evaluate our model's ability to forecast fashion.
We quantify its performance against an array of alternative models, both in terms of forecasters and alternative representations.
We also demonstrate its potential power for providing interpretable forecasts, analyzing style dynamics, and forecasting individual fashion elements. 

\begin{figure}[t]
\centering
    \includegraphics[width=0.85\linewidth]{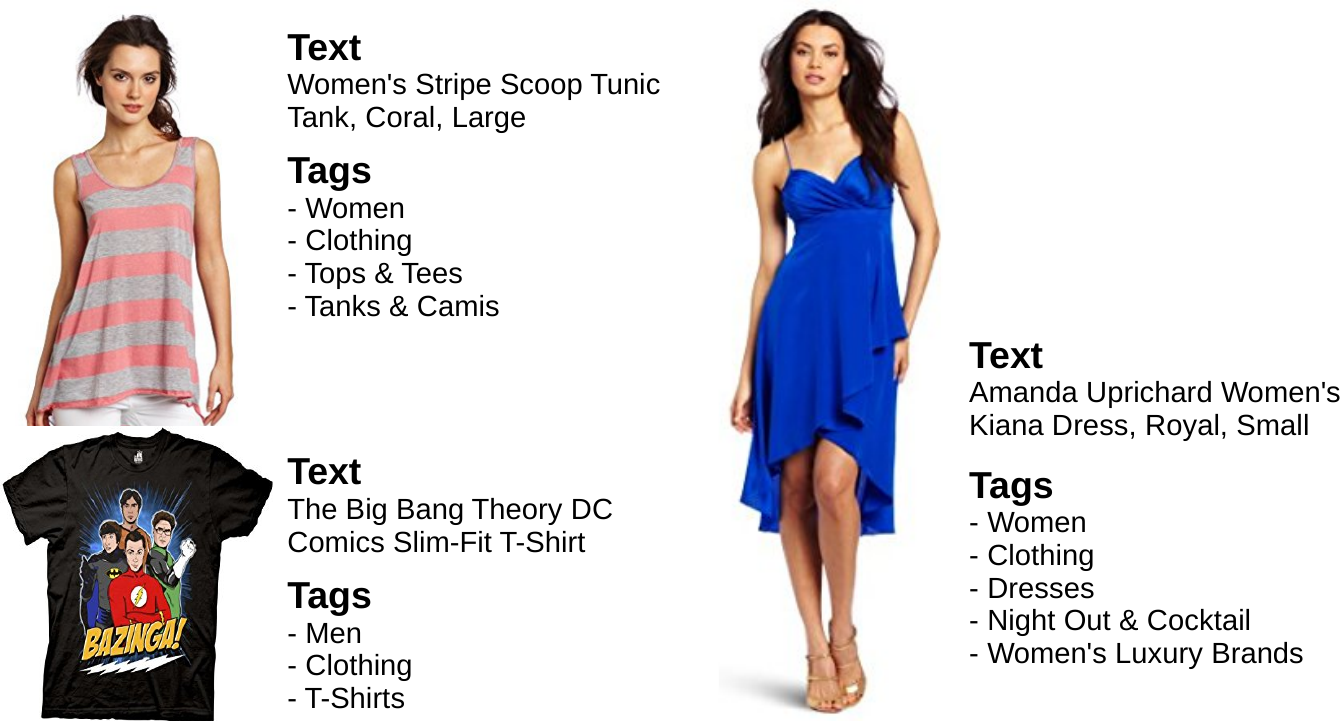}
\figlblvspace
\caption{The fashion items are represented with an image, a textual description, and a set of tags.}
\label{fig:data_sample}
\figvspace
\end{figure}

\paragraph{Datasets}
We evaluate our approach on three datasets collected from \textit{Amazon} by \cite{McAuley2015}.
The datasets represent three garment categories for women (Dresses and Tops\&Tees) and men (Shirts).
An item in these sets is represented with a picture, a short textual description, and a set of tags (see \figref{fig:data_sample}).
Additionally, it contains the dates each time the item was purchased.

These datasets are a good testbed for our model since they capture real-world customers' preferences in fashion and they span a fairly long period of time.
For all experiments, we consider the data in the time range from January 2008 to December 2013.
We use the data from the years 2008 to 2011 for training, 2012 for validation, and 2013 for testing.
\tblref{tab:datasets} summarizes the dataset sizes.

\begin{table}[t]
\centering
\scalebox{0.8}{
\begin{tabular}{l c c}

\toprule
Dataset         & \#Items & \#Transaction\\
\midrule
Dresses 		& 19,582 & 55,956 \\
Tops \& Tees	& 26,848 & 67,338 \\
Shirts 			& 31,594 & 94,251 \\

\bottomrule

\end{tabular}
}
\tablblvspace
\caption{Statistics of the three datasets from Amazon.}
\label{tab:datasets}
\tabvspace
\end{table}  
  \subsection{Style discovery}\label{sec:eval_style}
We use our deep model trained on DeepFashion~\cite{Liu2016} (cf.~\secref{sec:app_attributes}) to infer the semantic attributes for all items in the three datasets, and then learn $K=30$ styles from each.
We found that learning around 30 styles within each category is sufficient to discover interesting visual styles that are not too generic with large within-style variance nor too specific, \ie, describing only few items in our data. 
Our attribute predictions average 83\% AUC 
on a held-out DeepFashion validation set;
attribute ground truth is unavailable for the Amazon datasets themselves.

\figref{fig:styles_topimgs} shows 15 of the discovered styles in 2 of the datasets along with the 3 top ranked items based on the likelihood of that style in the items $p(s_k|x_i)$, and the most likely attributes per style ($p(a_m|s_k)$).
As anticipated, our model automatically finds the fine-grained styles within each genre of clothing.
While some styles vary across certain dimensions, there is a certain set of attributes that identify the style signature.
For example, color is not a significant factor in the $1^{st}$ and $3^{rd}$ styles (indexed from left to right) of Dresses. 
It is the mixture of shape, design, and structure that defines these styles (\emph{sheath}, \emph{sleeveless} and \emph{bodycon} in $1^{st}$, and \emph{chiffon}, \emph{maxi} and \emph{pleated} in $3^{rd}$).
On the other hand, the clothing material might dominate certain styles, like \emph{leather} and \emph{denim} in the $11^{th}$ and $15^{th}$  style of Dresses.  
Having a Dirichlet prior for the style distribution over the attributes induces sparsity. 
Hence, our model focuses on the most distinctive attributes for each style.
A naive approach (\eg, clustering) could be distracted by the many visual factors and become biased towards certain properties like color, \eg, by grouping all black clothes in one style while ignoring subtle differences in shape and material.
 \newcommand{\flws}{0.88}
\begin{figure*}[t]
\centering
\begin{subfigure}[t]{\flws\linewidth}
    \centering
    \includegraphics[width=\linewidth]{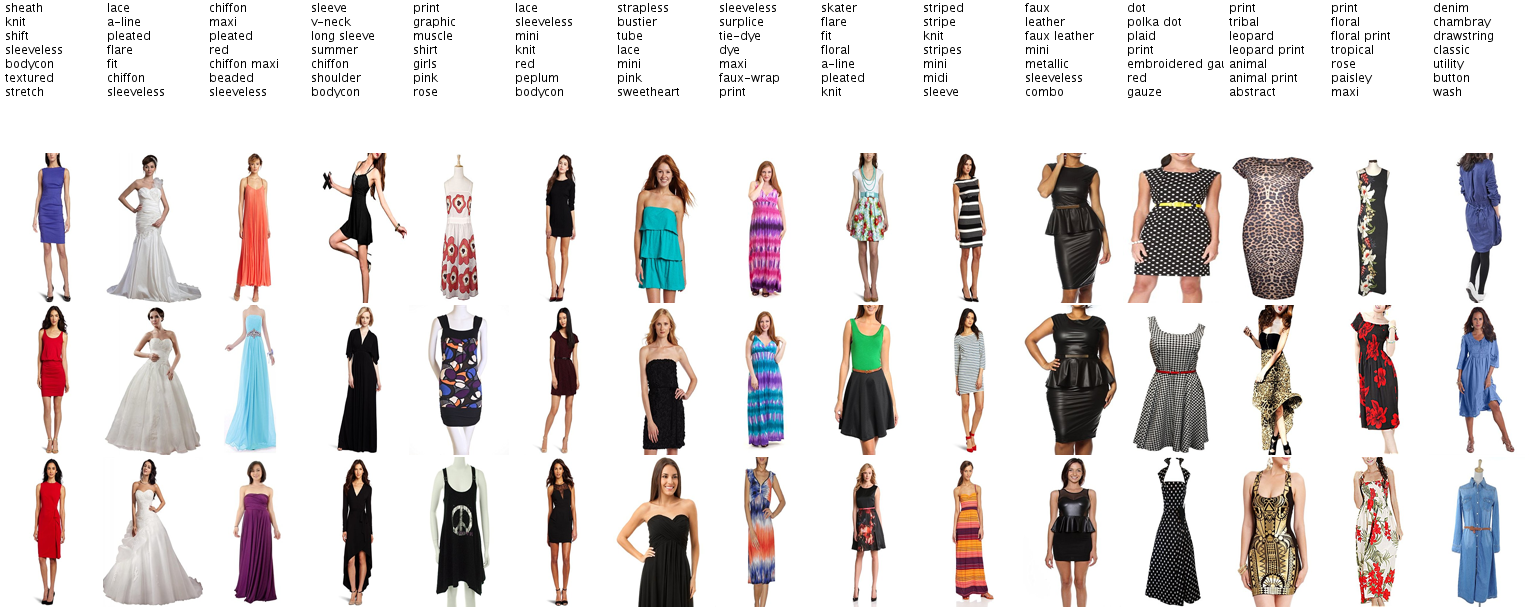}
    \caption{Dresses}\label{fig:style_dresses}
\end{subfigure}\\
\vspace{2mm}
\begin{subfigure}[t]{\flws\linewidth}
    \centering
    \includegraphics[width=\linewidth]{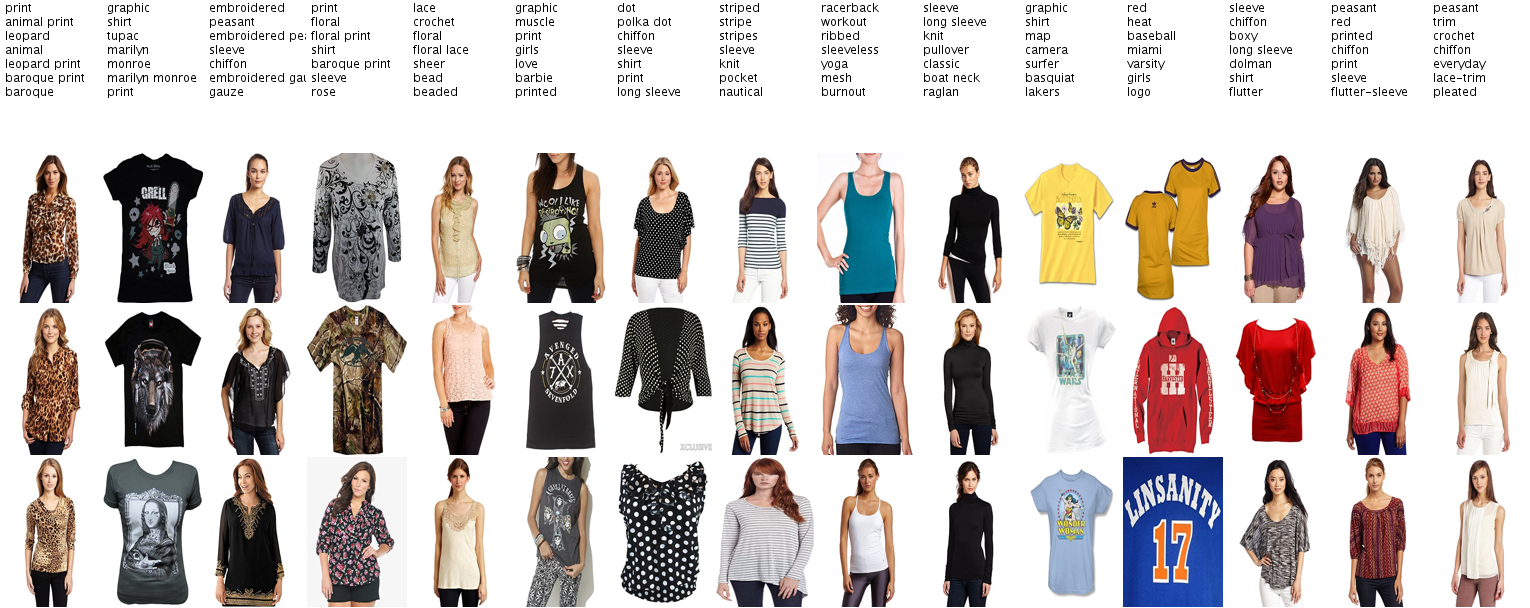}
    \caption{Tops \& Tees}\label{fig:style_topstees}
\end{subfigure}\\
\figlblvspace
\caption{The discovered visual styles on (a) Dresses and (b) Tops \& Tees datasets (see Supp for Shirts). Our model captures the fine-grained differences among the styles within each genre and provides a semantic description of the style signature based on visual attributes.}
\label{fig:styles_topimgs}
\figvspace
\end{figure*}
 \subsection{Style forecasting}\label{sec:eval_forecast}
Having discovered the latent styles in our datasets, we construct their temporal trajectories as in \secref{sec:app_forecast} using a temporal resolution of months.
We compare our approach to several well-established forecasting baselines, which we group in three main categories:

\paragraph{Na\"{i}ve}
These methods rely on the general properties of the trajectory: 
\begin{enumerate*}[label={\arabic*)}, itemjoin={;~}]
\item \emph{mean}: it forecasts the future values to be equal to the mean of the observed series
\item \emph{last}: it assumes the forecast to be equal to the last observed value
\item \emph{drift}: it considers the general trend of the series.
\end{enumerate*}

\paragraph{Autoregression}
These are linear regressors based on the last few observed values' ``lags''.
We consider several variations~\cite{box2015time}:
\begin{enumerate*}[label={\arabic*)}, itemjoin={;~}]
\item The linear autoregression model (\emph{AR})
\item the AR model that accounts for seasonality (\emph{AR+S})
\item the vector autoregression (\emph{VAR}) that considers the correlations between the different styles' trajectories
\item and the autoregressive integrated moving average model \emph(ARIMA).
\end{enumerate*}

\paragraph{Neural Networks}
Similar to autoregression, the neural models rely on the previous lags to predict the future; however these models incorporate nonlinearity which make them more suitable to model complex time series.
We consider two architectures with sigmoid non-linearity:
\begin{enumerate*}[label={\arabic*)}, itemjoin={;~}]
\item The feed forward neural network (\emph{FFNN})
\item and the time lagged neural network (\emph{TLNN}) \cite{faraway1998time}.
\end{enumerate*}

For models that require stationarity (\eg AR), we consider the differencing order as a hyperparamtere for each style.
All hyperparameters ($\alpha$ for ours, number of lags for the autoregression, and hidden neurons for neural networks) are estimated over the validation split of the dataset.
We compare the models based on two metrics: The mean absolute error $\mathrm{MAE} = \frac{1}{n}\sum_{t=1}^{n}|e_t|$, and the mean absolute percentage error $\mathrm{MAPE} = \frac{1}{n}\sum_{t=1}^{n}|\frac{e_t}{y_t}| \times 100$.
Where $e_t=\hat{y}_t-y_t$ is the error in predicting $y_t$ with $\hat{y}_t$.

\paragraph{Forecasting results}
\tblref{tab:forecast_perf} shows the forecasting performance of all models on the test data.
Here, all models use the identical visual style representation, namely our attribute-based NMF approach.
Our exponential smoothing model outperforms all baselines across the three datasets.
Interestingly, the more involved models like ARIMA, and the neural networks do not perform better.
This may be due to their larger number of parameters and the relatively short style trajectories.
Additionally, no strong correlations among the styles were detected and VAR showed inferior performance.
We expect there would be higher influence between styles from different garment categories rather than between styles within a category.
Furthermore, modeling seasonality (AR+S) does not improve the performance of the linear autoregression model.
We notice that the Dresses dataset is more challenging than the other two.
The styles there exhibit more temporal variations compared to the ones in Tops\&Tees and Shirts, which may explain the larger forecast error in general.
Nonetheless, our model generates a reliable forecast of the popularity of the styles for a year ahead across all data sets.
The forecasted style trajectory by our approach is within a close range to the actual one (only 3 to 6 percentage error based on MAPE).
Furthermore, we notice that our model is not very sensitive to the number of styles.
When varying K between 15 and 85, the relative performance of the forecast approaches is similar to \tblref{tab:forecast_perf}, with EXP performing the best.

\begin{figure}[t]
\centering
\includegraphics[width=0.98\linewidth]{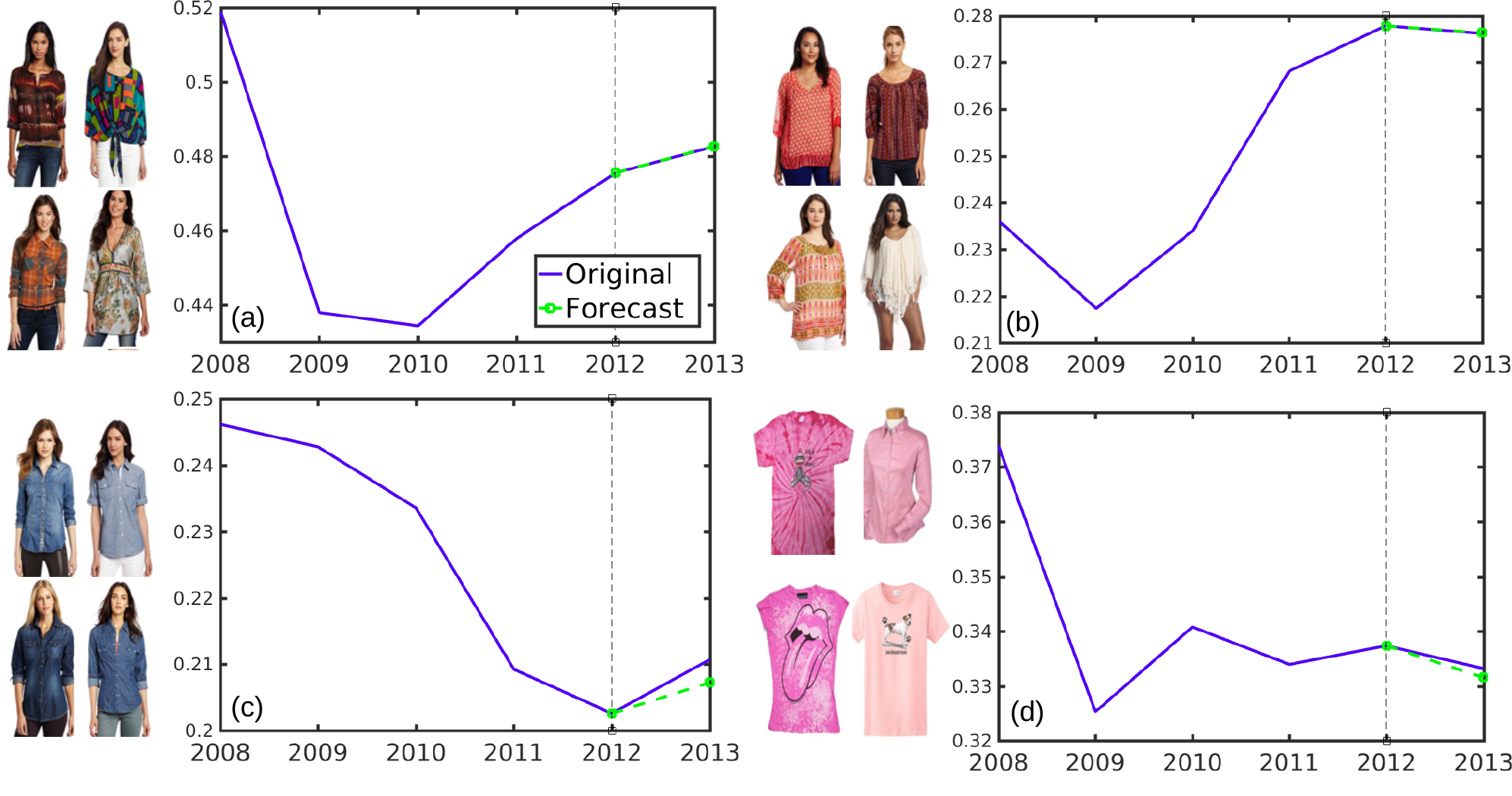}
\figlblvspace
\caption{The forecasted popularity estimated by our model for 4 styles from the Tops \& Tees dataset. Our model successfully predicts the popularity of styles in the future and performs well even with challenging trajectories that experience a sudden change in direction like in (c) and (d).}
\label{fig:forecast}
\figvspace
\end{figure}

\figref{fig:forecast} visualizes our model's predictions on four styles from the Tops\&Tees dataset.
For trajectories in \figref{fig:forecast}a and \figref{fig:forecast}b, our approach successfully captures the popularity of styles in year 2013.
Styles in \figref{fig:forecast}c and \figref{fig:forecast}d are much more challenging.
Both of them experience a reflection point at year 2012, from a declining popularity to an increase and vice versa.
Still, the predictions made by our model forecast this change in direction correctly and the error in the estimated popularity is minor.
 
\begin{table}[t]
\centering
\scalebox{0.8}{
\begin{tabular}{l c c c c c c}

\toprule
\multirow{ 2}{*}{Model}   &   \multicolumn{2}{c}{Dresses}   &  \multicolumn{2}{c}{Tops \& Tees}  &  \multicolumn{2}{c}{Shirts}   \\
               &    MAE  &   MAPE  &      MAE  &   MAPE   &    MAE  &   MAPE  \\
\midrule
\multicolumn{7}{l}{\textbf{Na\"{i}ve}} \\
        mean   &  0.0345  &  25.50  &   0.0513  &  17.61  &   0.0155  &  6.14   \\
        last   &  0.0192  &  8.38   &   0.0237  &  8.66   &   0.0160  &  5.50   \\
       drift   &  0.0201  &  9.17   &   0.0158  &  5.70   &   0.0177  &  6.50   \\
\midrule
\multicolumn{7}{l}{\textbf{Autoregression}} \\
          AR   &  0.0174  &  9.65   &   0.0148  &  \textbf{5.20}   &   0.0120  &  4.45   \\
        AR+S   &  0.0210  &  12.78  &   0.0177  &  6.41   &   0.0122  &  4.51   \\
         VAR   &  0.0290  &  20.36  &   0.0422  &  14.61  &   0.0150  &  5.92   \\
       ARIMA   &  0.0186  &  13.04  &   0.0154  &  5.45   &   0.0092  &  3.41   \\
\midrule
\multicolumn{7}{l}{\textbf{Neural Network}} \\
        TLNN   &  0.0833  &  35.45  &   0.0247  &  8.49   &   0.0124  &  4.24   \\
        FFNN   &  0.0973  &  41.18  &   0.0294  &  10.26  &   0.0109  &  3.97   \\
\midrule
      \ours{}  &     \textbf{0.0146}  &  \textbf{6.54}   &  \textbf{0.0145}  &  5.36   &   \textbf{0.0088}  &  \textbf{3.16}   \\
\bottomrule

\end{tabular}
}
\tablblvspace
\caption{The forecast error of our approach compared to several baselines on three datasets.}
\label{tab:forecast_perf}
\tabvspace
\end{table} 
  
  \subsection{Fashion representation}\label{sec:eval_rep}
Thus far we have shown the styles discovered by our approach as well as our ability to forecast the popularity of visual styles in the future.
Next we examine the impact of our representation compared to both textual meta-data and CNN-based alternatives.

\paragraph{Meta Information}
Fashion items are often accompanied by information other than the images.
We consider two types of meta information supplied with the Amazon datasets (\figref{fig:data_sample}):
\begin{enumerate*}[label={\arabic*)}, itemjoin={;~}]
\item \emph{Tags}: which identify the categories, the age range, the trademark, the event, \etc.
\item \emph{Text}: which provides a description of the item in natural language.
\end{enumerate*}
For both, we learn a unique vocabulary of tags and words across the dataset and represent each item using a bag of words representation.
From thereafter, we can employ our NMF and forecasting models just as we do with our visual attribute-based vocabulary.
In results, we consider a text-only baseline as well as a multi-modal approach that augments our attribute model with textual cues.

\paragraph{Visual}
Attributes are attractive in this problem setting for their interpretability, but how fully do they capture the visual content?
To analyze this, we implement an alternative representation based on deep features extracted from a pre-trained convolutional neural network (CNN).
In particular, we train a CNN with an AlexNet-like architecture on the DeepFashion dataset to perform \emph{clothing classification} (see Supp.~for details).
Since fashion elements can be local properties (\eg, v-neck) or global (\eg, a-line), we use the CNN to extract two representations at different abstraction levels:
\begin{enumerate*}[label={\arabic*)}, itemjoin={;~}]
\item \emph{FC7}: features extracted from the last hidden layer
\item \emph{M3}: features extracted from the third max pooling layer after the last convolutional layer.
\end{enumerate*}
We refer to these as ClothingNet-FC7 and ClothingNet-M3 in the following.

\paragraph{Forecasting results}
The textual and visual cues inherently rely on distinct vocabularies, and the metrics applied for \tblref{tab:forecast_perf} are not comparable across representations.
Nonetheless, we can gauge their relative success in forecasting by measuring the distribution difference between their predictions and the ground truth styles, in their respective feature spaces.
In particular, we apply the experimental setup of \secref{sec:eval_forecast}, then record the Kullback-Leibler divergences (KL) between the forecasted distribution and the actual test set distribution.
For all models, we apply our best performing forecaster from \tblref{tab:forecast_perf} (EXP).  

\tblref{tab:representation_perf_scaled} shows the effect of each representation on forecasting across all three datasets.
Among all single modality methods, ours is the best.
Compared to the ClothingNet CNN baselines, our attribute styles are much more reliable.
Upon visual inspection of the learned styles from the CNNs, we find out that they are sensitive to the pose and spatial configuration of the item and the person in the image.
This reduces the quality of the discovered styles and introduces more noise in their trajectories.
Compared to the tags alone, the textual description is better, likely because it captures more details about the appearance of the item.
However, compared to any baseline based only on meta data, our approach is best.
This is an important finding: \emph{predicted} visual attributes yield more reliable fashion forecasting than strong real-world meta-data cues.
To see the future of fashion, it pays off to really look at the images themselves.

The bottom of \tblref{tab:representation_perf_scaled} shows the results when 
using various combinations of text and tags along with attributes.
We see that our model is even stronger, arguing for including meta-data with visual data whenever it is available.

\begin{table}[t]
\centering
\scalebox{0.70}{
\begin{tabular}{l c c c c c c}

\toprule
\multirow{ 2}{*}{Model} & \multicolumn{2}{c}{Dresses} 			& \multicolumn{2}{c}{Tops \& Tees} 	& \multicolumn{2}{c}{Shirts} \\
						& \multicolumn{1}{c}{KL} 	& 	IMP(\%) & \multicolumn{1}{c}{KL} & 	IMP(\%) & \multicolumn{1}{c}{KL} & 	IMP(\%) \\

\midrule
\multicolumn{7}{l}{\textbf{Meta Information}}\\
Tags 	 		&  0.0261 	 &  0  	 	 &  0.0161 	 &  0	  	 &  0.0093 	 &  0	    \\
Text 	 		&  0.0185 	 &  29.1  	 &  0.0075 	 &  53.4  	 &  0.0055 	 &  40.9    \\
\multicolumn{7}{l}{\textbf{Visual}}\\
ClothingNet-FC7 &  0.0752 	 &  -188.1   &  0.25 	 &  -1452.8  &  0.1077 	 &  -1058.1 \\
ClothingNet-M3 	&  0.0625 	 &  -139.5   &  0.0518 	 &  -221.7   & 	0.0177 	 & 	-90.3   \\
Attributes 		&  \textbf{0.0105} 	 &  \textbf{59.8}  	 &  \textbf{0.0049} 	 &  \textbf{69.6}  	 &  \textbf{0.0035} 	 &  \textbf{62.4}    \\
\midrule
\multicolumn{7}{l}{\textbf{Multi-Modal}}\\
Attributes+Tags &  0.0336 	 &  -28.7  	 &  0.0099 	 &  38.5  	 &  0.0068 	 &  26.9    \\
Attributes+Text &  0.0051 	 &  80.5  	 &  0.0053 	 &  67.1  	 &  \textbf{0.0014} 	 &  \textbf{84.9}    \\
Attr+Tags+Text 	&  \textbf{0.0041} 	 &  \textbf{84.3}  	 &  \textbf{0.0052} 	 &  \textbf{67.7}  	 &  \textbf{0.0014} 	 &  \textbf{84.9}    \\

\bottomrule

\end{tabular}
}
\tablblvspace
\caption{Forecast performance for various fashion representations in terms of KL divergence (lower is better) and the relative improvement (IMP) over the Tags baseline (higher is better).
Our attribute-based visual styles lead to much more reliable forecasts compared to meta data or other visual representations.
}
\label{tab:representation_perf_scaled}
\tabvspace
\end{table}
 \subsection{Style dynamics}
\begin{figure}[t!]
\centering
\includegraphics[width=\linewidth]{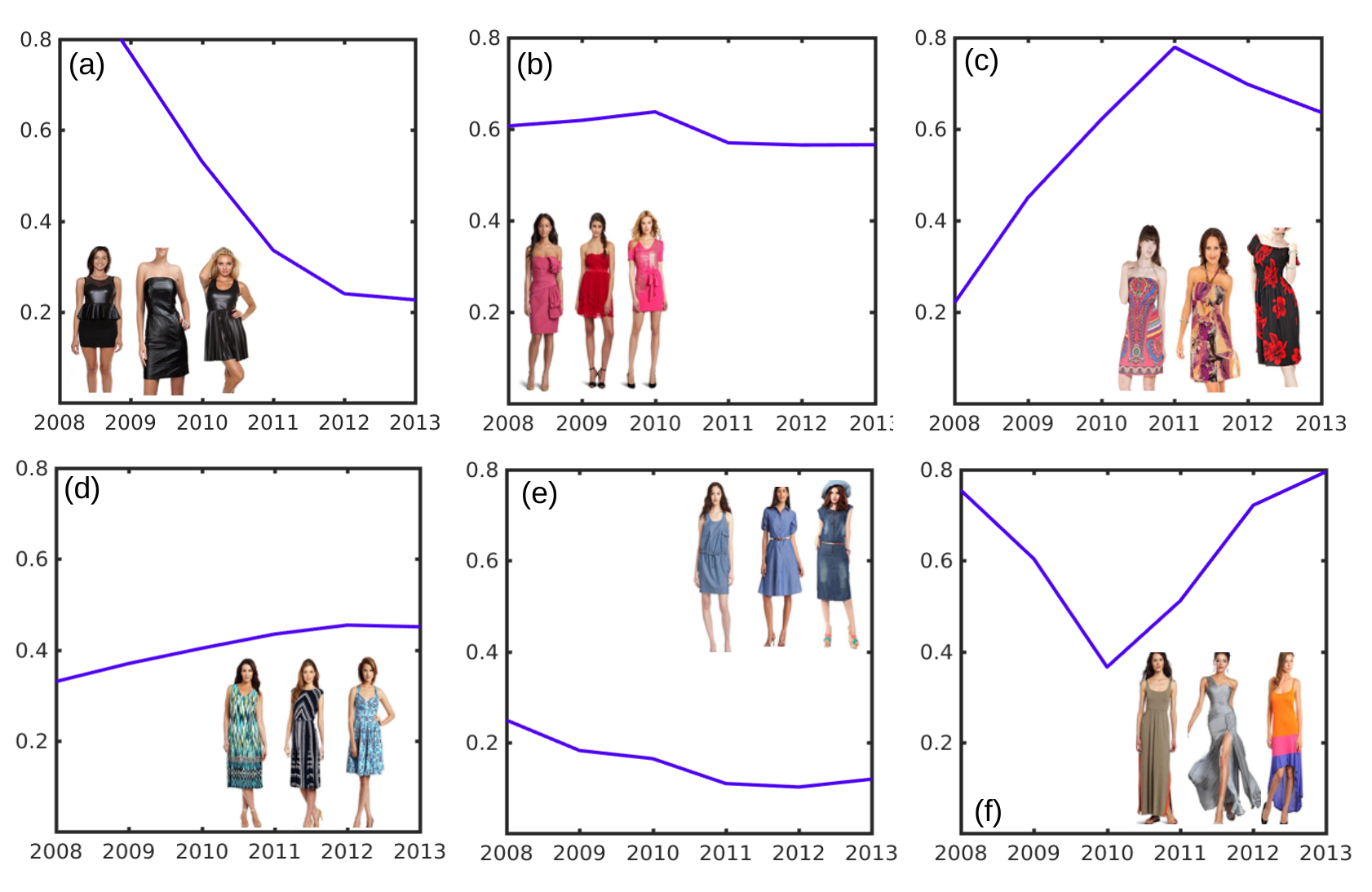}
\figlblvspace
\caption{Our approach offers the unique opportunity to examine the life cycle of visual styles in fashion. Some interesting temporal dynamics of the styles discovered by our model can be grouped into: (a) out of fashion; (b) classic; (c) in fashion or (d) trending; (e)  unpopular; and (f) re-emerging styles.
}
\label{fig:yearly_trend}
\figvspace
\end{figure}

Having established the ability to forecast visual fashions, we now turn to demonstrating some suggestive applications.
Fashion is a very active domain with styles and designs going in and out of popularity at varying speeds and stages.
The life cycle of fashion goes through four main stages \cite{sproles1981analyzing}: 1) introduction; 2) growth; 3) maturity; and finally 4) decline.
Knowing which style is at which level of its lifespan is of extreme importance for the fashion industry.
Understanding the style dynamics helps companies to adapt their strategies and respond in time to accommodate the customers' needs.
Our model offers the opportunity to inspect visual style trends and lifespans.
In \figref{fig:yearly_trend}, we visualize the temporal trajectories computed by our model for 6 styles from Dresses.
The trends reveal several categories of styles:
\begin{enumerate*}[label={\arabic*)}, itemjoin={;~}]
	\item \emph{Out of fashion}: styles that are losing popularity at a rapid rate (\figref{fig:yearly_trend}a)
	\item \emph{Classic}: styles that are relatively popular and show little variations through the years (\figref{fig:yearly_trend}b)
	\item \emph{Trending}: styles that are trending and gaining popularity at a high rate (\figref{fig:yearly_trend}c and d)
	\item \emph{Unpopular}: styles that are currently at a low popularity rate with no sign of improvement (\figref{fig:yearly_trend}e)
	\item \emph{Re-emerging}: styles that were popular in the past, declined, and then resurface again and start trending (\figref{fig:yearly_trend}f).
\end{enumerate*}

Our model is in a unique position to offer this view point on fashion.
For example, using item popularity and trajectories is not informative about the life cycle of the visual style.
An item lifespan is influenced by many other factors such as pricing, marketing strategy, and advertising among many others.
By learning the latent visual styles in fashion, our model is able to capture the collective styles shared by many articles and, hence, depicts a more realistic popularity trajectory that is less influenced by irregularities experienced by the individual items.

 \subsection{Forecasting elements of fashion}

While so far we focused on visual style forecasting, our model is capable of inferring the popularity of the individual attributes as well.
Thus it can answer questions like: what kind of fabric, texture, or color will be popular next year?  These questions are of significant interest in the fashion industry (e.g., see the ``fashion oracle" World Global Style Network~\cite{wgsn,wgsn2}, which thousands of designers rely on for trend prediction on silhouettes, palettes, etc.).

We get the attribute popularity $p(a_m|t)$ at a certain time $t$ in the future through the forecasted popularity of the styles:
\begin{equation}
\eqtopvspace
p(a_m|t) = \sum_{s_k \in S} p(a_m|s_k)p(s_k|t)
\eqbottomvspace
\end{equation}
where $p(a_m|s_k)$ is the probability of attribute $a_m$ given style $s_k$ based on our style discovery model, and $p(s_k|t)$ is the forecated probability of style $s_k$ at time $t$.

For the 1000 attributes in our visual vocabulary, our model achieves an intersection with ground truth popularity rank at 90\%, 84\% and 88\% for the Top 10, 25 and 50 attributes respectively.
\figref{fig:attributes_cloud} shows the forecasted \emph{texture} and \emph{shape} attributes for the Dresses test set.
Our model successfully captures the most dominant attributes in both groups of attributes, correctly giving the gist of future styles.
 \newcommand{\flwa}{.48}
\begin{figure}[t]
\centering
\begin{subfigure}[t]{\flwa\linewidth}
    \centering
    \includegraphics[width=0.48\linewidth]{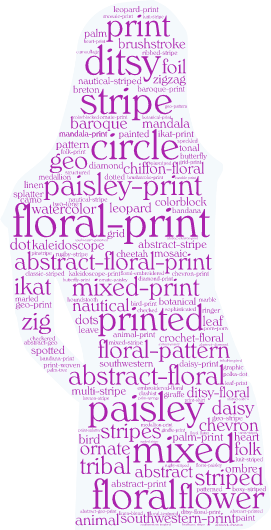}    
    \includegraphics[width=0.48\linewidth]{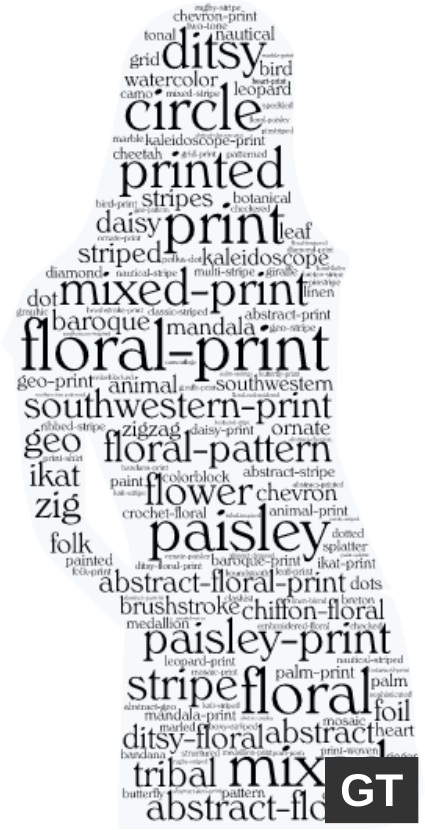}    
    \caption{Texture}\label{fig:texture}
\end{subfigure}
\begin{subfigure}[t]{\flwa\linewidth}
    \centering
    \includegraphics[width=0.48\linewidth]{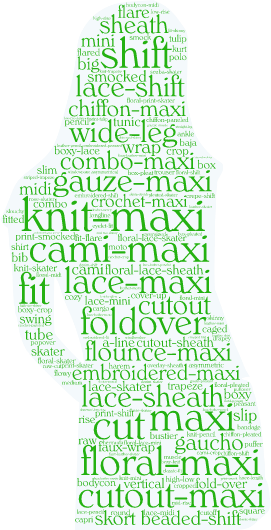}    
    \includegraphics[width=0.48\linewidth]{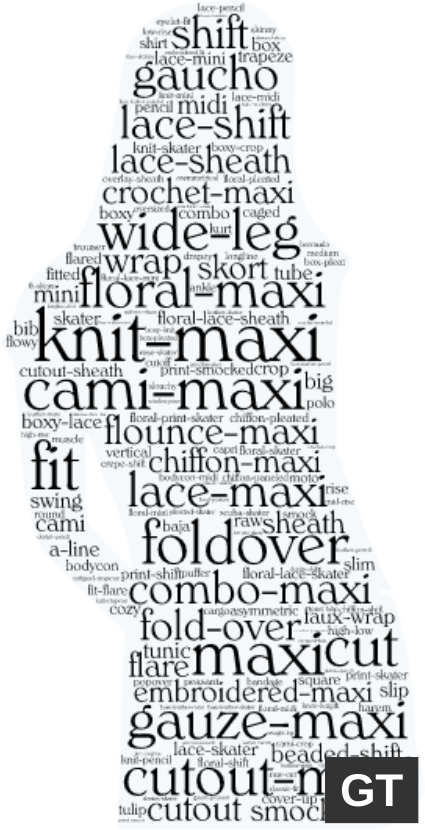}
    \caption{Shape}\label{fig:shape}
\end{subfigure}
\figlblvspace
\caption{Our model can predict the popularity of individual fashion attributes using the forecasted styles as a proxy. The forecasted attributes are shown in color while the ground truth is in black. The attribute size is relative to its popularity rank.}
\label{fig:attributes_cloud}
\figvspace
\end{figure} 
 \section{Conclusion}
In the fashion industry, predicting trends, due to its complexity, is frequently compared to weather forecasting: sometimes you get it right and sometimes you get it wrong.
In this work, we show that using our vision-based fashion forecasting model we get it right more often than not. 
We propose a model that discovers fine-grained visual styles from large scale fashion data in an unsupervised manner.
Our model identifies unique style signatures and provides a semantic description for each based on key visual attributes.
Furthermore, based on user consumption behavior, our model predicts the future popularity of the styles, and reveals their life cycle and status (\eg in- or out of fashion).
We show that vision is essential for reliable forecasts, outperforming textual-based representations.
Finally, fashion is not restricted to apparel; it is present in accessories, automobiles, and even house furniture.
Our model is generic enough to be employed in different domains where a notion of visual style is present.
 
\balance
{\small
\bibliographystyle{ieee}
\bibliography{mybib}

\begin{thebibliography}{10}\itemsep=-1pt

\bibitem{fashionunited}
{Fashion Statistics}.
\newblock \url{https://fashionunited.com/global-fashion-industry-statistics}.

\bibitem{wgsn2}
{Trend-Forecasting}.
\newblock
  \url{http://fusion.net/story/305446/wgsn-trend-forecasting-sarah-owen/}.

\bibitem{wgsn}
{WGSN}.
\newblock \url{https://www.wgsn.com/en/}.

\bibitem{Berg2010}
T.~L. Berg, A.~C. Berg, and J.~Shih.
\newblock {Automatic Attribute Discovery and Characterization from Noisy Web
  Data}.
\newblock In {\em ECCV}, 2010.

\bibitem{Bossard2012}
L.~Bossard, M.~Dantone, C.~Leistner, C.~Wengert, T.~Quack, and L.~{Van Gool}.
\newblock {Apparel Classification with Style}.
\newblock In {\em ACCV}, 2012.

\bibitem{box2015time}
G.~E. Box, G.~M. Jenkins, G.~C. Reinsel, and G.~M. Ljung.
\newblock {\em Time series analysis: forecasting and control}.
\newblock John Wiley \& Sons, 2015.

\bibitem{Bracher2016}
C.~Bracher, S.~Heinz, and R.~Vollgraf.
\newblock {Fashion DNA: Merging Content and Sales Data for Recommendation and
  Article Mapping}.
\newblock In {\em KDD Fashion Workshop}, 2016.

\bibitem{brown1961fundamental}
R.~G. Brown and R.~F. Meyer.
\newblock The fundamental theorem of exponential smoothing.
\newblock {\em Operations Research}, 9(5):673--685, 1961.

\bibitem{Chen2012}
H.~Chen, A.~Gallagher, and B.~Girod.
\newblock {Describing Clothing by Semantic Attributes}.
\newblock In {\em ECCV}, 2012.

\bibitem{Chen2015Devils}
K.~Chen, K.~Chen, P.~Cong, W.~H. Hsu, and J.~Luo.
\newblock Who are the devils wearing prada in new york city?
\newblock In {\em ACM Multimedia}, 2015.

\bibitem{Chen2015DeepDomain}
Q.~Chen, J.~Huang, R.~Feris, L.~M. Brown, J.~Dong, and S.~Yan.
\newblock {Deep Domain Adaptation for Describing People Based on Fine-Grained
  Clothing Attributes}.
\newblock In {\em CVPR}, 2015.

\bibitem{DavisBook}
K.~Davis.
\newblock {\em Don't Know Much About History: Everything You Need to Know About
  American History but Never Learned}.
\newblock Harper, 2012.

\bibitem{Di2013}
W.~Di, C.~Wah, A.~Bhardwaj, R.~Piramuthu, and N.~Sundaresan.
\newblock {Style finder: Fine-grained clothing style detection and retrieval}.
\newblock In {\em CVPR Workshops}, 2013.

\bibitem{faraway1998time}
J.~Faraway and C.~Chatfield.
\newblock Time series forecasting with neural networks: a comparative study
  using the airline data.
\newblock {\em Applied statistics}, pages 231--250, 1998.

\bibitem{He2016}
R.~He and J.~McAuley.
\newblock {Ups and Downs: Modeling the Visual Evolution of Fashion Trends with
  One-Class Collaborative Filtering}.
\newblock In {\em WWW}, 2016.

\bibitem{Hsiao2017}
W.-L. Hsiao and K.~Grauman.
\newblock {Learning the Latent ``Look'': Unsupervised Discovery of a
  Style-Coherent Embedding from Fashion Images}.
\newblock In {\em ICCV}, 2017.

\bibitem{Hu2015NTF}
C.~Hu, P.~Rai, C.~Chen, M.~Harding, and L.~Carin.
\newblock {Scalable Bayesian Non-Negative Tensor Factorization for Massive
  Count Data}.
\newblock In {\em ECML PKDD}, 2015.

\bibitem{Huang2015}
J.~Huang, R.~Feris, Q.~Chen, and S.~Yan.
\newblock {Cross-Domain Image Retrieval With a Dual Attribute-Aware Ranking
  Network}.
\newblock In {\em ICCV}, 2015.

\bibitem{Iwata2011}
T.~Iwata, S.~Watanabe, and H.~Sawada.
\newblock {Fashion Coordinates Recommender System Using Photographs from
  Fashion Magazines}.
\newblock In {\em IJCAI International Joint Conference on Artificial
  Intelligence}, 2011.

\bibitem{Kiapour2015}
M.~H. Kiapour, X.~Han, S.~Lazebnik, A.~C. Berg, and T.~L. Berg.
\newblock {Where to Buy It: Matching Street Clothing Photos in Online Shops}.
\newblock In {\em ICCV}, 2015.

\bibitem{Kiapour2014}
M.~H. Kiapour, K.~Yamaguchi, A.~C. Berg, and T.~L. Berg.
\newblock {Hipster wars: Discovering Elements of Fashion Styles}.
\newblock In {\em ECCV}, 2014.

\bibitem{Kingma2015}
D.~P. Kingma and J.~L. Ba.
\newblock {ADAM: A Method for Stochastic Optimization}.
\newblock In {\em ICLR}, 2015.

\bibitem{kolda2009tensor}
T.~G. Kolda and B.~W. Bader.
\newblock Tensor decompositions and applications.
\newblock {\em SIAM review}, 51(3):455--500, 2009.

\bibitem{Kovashka2012}
A.~Kovashka, D.~Parikh, and K.~Grauman.
\newblock {WhittleSearch: Image search with relative attribute feedback}.
\newblock In {\em CVPR}, 2012.

\bibitem{Krizhevsky2012}
A.~Krizhevsky, I.~Sutskever, and G.~E. Hinton.
\newblock {ImageNet Classification with Deep Convolutional Neural Networks}.
\newblock In {\em NIPS}, 2012.

\bibitem{Kwak2013}
I.~Kwak, A.~Murillo, P.~Belhumeur, D.~Kriegman, and S.~Belongie.
\newblock {From Bikers to Surfers: Visual Recognition of Urban Tribes}.
\newblock In {\em BMVC}, 2013.

\bibitem{Liu2012Magic}
S.~Liu, J.~Feng, Z.~Song, T.~Zhang, H.~Lu, C.~Xu, and S.~Yan.
\newblock {``Hi, Magic Closet, Tell Me What to Wear !''}.
\newblock In {\em ACM Multimedia}, 2012.

\bibitem{Liu2012Street}
S.~Liu, Z.~Song, G.~Liu, C.~Xu, H.~Lu, and S.~Yan.
\newblock {Street-to-shop: Cross-scenario clothing retrieval via parts
  alignment and auxiliary set}.
\newblock In {\em CVPR}, 2012.

\bibitem{Liu2016}
Z.~Liu, S.~Qiu, and X.~Wang.
\newblock {DeepFashion : Powering Robust Clothes Recognition and Retrieval with
  Rich Annotations}.
\newblock In {\em CVPR}, 2016.

\bibitem{McAuley2015}
J.~McAuley, C.~Targett, Q.~Shi, and A.~van~den Hengel.
\newblock {Image-based Recommendations on Styles and Substitutes}.
\newblock In {\em ACM SIGIR}, 2015.

\bibitem{Murillo2012}
A.~C. Murillo, I.~S. Kwak, L.~Bourdev, D.~Kriegman, and S.~Belongie.
\newblock {Urban tribes: Analyzing group photos from a social perspective}.
\newblock In {\em CVPR Workshops}, 2012.

\bibitem{Parikh2011}
D.~Parikh and K.~Grauman.
\newblock {Relative Attributes}.
\newblock In {\em ICCV}, 2011.

\bibitem{Simo-Serra2015}
E.~Simo-Serra, S.~Fidler, F.~Moreno-Noguer, and R.~Urtasun.
\newblock {Neuroaesthetics in Fashion: Modeling the Perception of
  Fashionability}.
\newblock In {\em CVPR}, 2015.

\bibitem{Simo-Serra2016}
E.~Simo-Serra and H.~Ishikawa.
\newblock {Fashion Style in 128 Floats : Joint Ranking and Classification using
  Weak Data for Feature Extraction}.
\newblock In {\em CVPR}, 2016.

\bibitem{Song2011}
Z.~Song, M.~Wang, X.-s. Hua, and S.~Yan.
\newblock {Predicting Occupation via Human Clothing and Contexts}.
\newblock In {\em ICCV}, 2011.

\bibitem{sproles1981analyzing}
G.~B. Sproles.
\newblock Analyzing fashion life cycles: principles and perspectives.
\newblock {\em The Journal of Marketing}, pages 116--124, 1981.

\bibitem{Vaccaro2016}
K.~Vaccaro, S.~Shivakumar, Z.~Ding, K.~Karahalios, and R.~Kumar.
\newblock {The Elements of Fashion Style}.
\newblock In {\em UIST}, 2016.

\bibitem{Veit2015}
A.~Veit, B.~Kovacs, S.~Bell, J.~McAuley, K.~Bala, and S.~Belongie.
\newblock {Learning Visual Clothing Style with Heterogeneous Dyadic
  Co-occurrences}.
\newblock In {\em ICCV}, 2015.

\bibitem{Vittayakorn2016}
S.~Vittayakorn, A.~C. Berg, and T.~L. Berg.
\newblock When was that made?
\newblock In {\em WACV}, 2017.

\bibitem{Vittayakorn2016a}
S.~Vittayakorn, T.~Umeda, K.~Murasaki, K.~Sudo, T.~Okatani, and K.~Yamaguchi.
\newblock {Automatic Attribute Discovery with Neural Activations}.
\newblock In {\em ECCV}, 2016.

\bibitem{Vittayakorn2015}
S.~Vittayakorn, K.~Yamaguchi, A.~C. Berg, and T.~L. Berg.
\newblock {Runway to realway: Visual analysis of fashion}.
\newblock In {\em WACV}, 2015.

\bibitem{paperdoll-iccv2013}
K.~Yamaguchi, H.~Kiapour, and T.~Berg.
\newblock Paper doll parsing: Retrieving similar styles to parse clothing
  items.
\newblock In {\em ICCV}, 2013.

\bibitem{yamaguchi-cvpr2012}
K.~Yamaguchi, H.~Kiapour, L.~Ortiz, and T.~Berg.
\newblock Parsing clothing in fashion photographs.
\newblock In {\em CVPR}, 2012.

\bibitem{Yu2015}
A.~Yu and K.~Grauman.
\newblock Just noticeable differences in visual attributes.
\newblock In {\em ICCV}, 2015.

\end{thebibliography}
}
\clearpage
\section{Appendix}
This appendix provides additional information for:
\begin{itemize}
	\item The deep attribute and the ClothingNet architectures.
	\item The forecast baseline models.
	\item The discovered topics on the Shirts dataset (see \figref{fig:styles_topimgs_supp}).
	\item Forecast examples of our model in comparison to the baselines on the three datasets (see \figref{fig:forecast_supp}).
\end{itemize}

\subsection{The deep attribute model}
\begin{figure}[t]
\centering
    \includegraphics[width=0.95\linewidth]{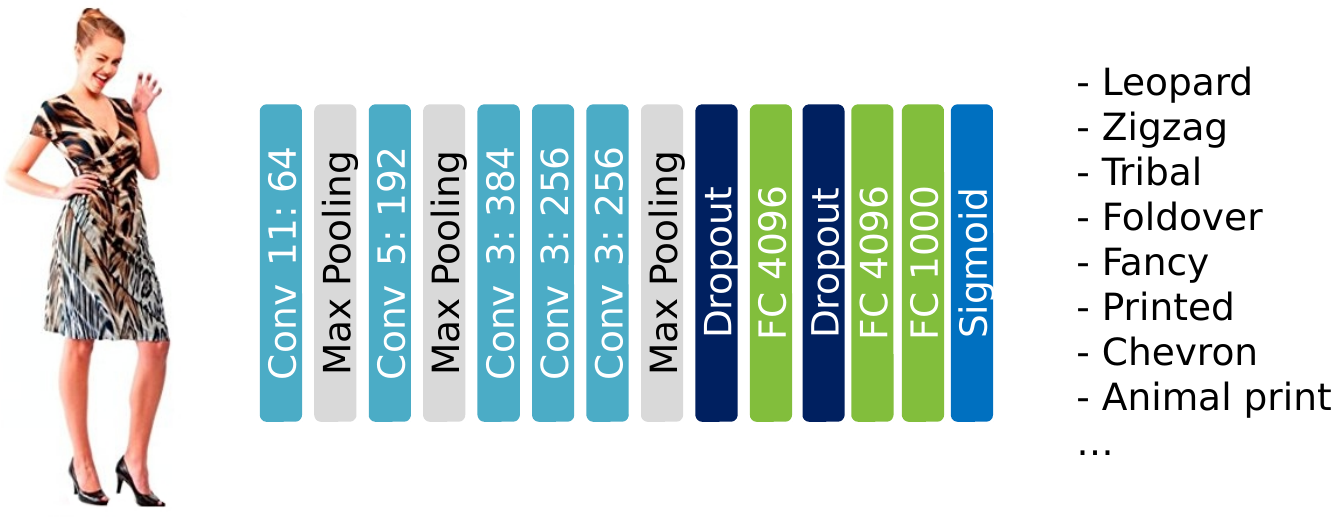}
\caption{The architecture of our deep attribute CNN model.}
\label{fig:attr_cnn_supp}
\end{figure}

\figref{fig:attr_cnn_supp} shows the details of the network architecture for our attribute prediction model.
The model is composed of 5 convolutional layers with decreasing filter sizes from $11\times 11$ to $3 \times 3$ followed by 3 fully connected layers and 2 dropout layers with probability of 0.5.
Additionally, each convolutional layer and the first two fully connected layers in our model are followed by a batch normalization layer and a rectified linear unit (ReLU).
For information on the training procedure and the hyperparameters see Section 3.1 in the main submission.

\subsection{ClothingNet}
The ClothingNet model is similar to our attribute model architecture with the last sigmoid layer replaced with a softmax.
The network is trained to distinguish 50 categories of garments (\eg \emph{Sweater}, \emph{Skirt}, \emph{Jeans} and \emph{Jacket}) from the DeepFashion dataset.
The model is trained for 45 epochs using Adam \cite{Kingma2015}.
On a held-out test set on DeepFashion, the ClothingNet achieves 86.5\% Top-5 accuracy.

\subsection{Forecast models}\vspace{+.2cm}
\paragraph{Na\"{i}ve}
which includes three simple models:
\begin{enumerate}[label={\arabic*)},itemsep=0pt,leftmargin=*] 
\item \emph{mean}: the future values are forecasted to be equal to the mean of the observed series, \ie $\hat{y}_{n+1|n}=\frac{1}{n}\sum_{t=1}^{n}{y_t}$.
\item \emph{last}: the forecast is equal to the last observed value, \ie $\hat{y}_{n+h|n}=y_n$.
\item \emph{drift}: the forecast follows the general trend of the series, \ie $\hat{y}_{n+h|n}=y_n + \frac{h}{n-1}(y_n-y_1)$ where $h$ is the forecast horizon.
\end{enumerate}

\paragraph{Autoregressors}
these linear regressors assume the current value to be a linear function of the last observed values ``lags'', \ie $\hat{y}_{n}=b+\sum_i^P\alpha_iy_{n-i} + \epsilon$ where $b$ is a constant, $\{\alpha_i\}$ are the lag coefficients, $P$ is the maximum lag (set by cross validation in our case) and $\epsilon$ an error term.
We consider several variations of the model \cite{box2015time}:
\begin{enumerate}[label={\arabic*)},itemsep=0pt,leftmargin=*] 
\item \emph{AR}: the autoregressor in its standard form. 
\item \emph{AR+S}: which further incorporates seasonality, \eg for a series with 12 months seasonality the model will also consider the lag at $n-12$ along with most recent lags to predict the current value.
\item \emph{VAR}: the vector autoregoressor considers the correlations between the different styles trajectories when predicting the future.
\item \emph{ARIMA}: the autoregressive integrated moving average model which models the temporal trajectory with two polynomials, one for autoregression and the other for the moving average. In addition it can handle non-stationary signals through differencing operations (integration).
\end{enumerate}

\paragraph{Neural Networks (NN)}
Similar to the autoregressor, the neural models rely on the previous lags to predict the current value of the signal; however these models incorporate nonlinearity which make them more suitable to model complex time series.
We consider two architectures with sigmoid non-linearity:
\begin{enumerate}[label={\arabic*)},itemsep=0pt,leftmargin=*] 
\item \emph{TLNN}: the time lagged neural network \cite{faraway1998time}.
\item \emph{FFNN}: the feed forward neural network.
\end{enumerate}

\newcommand{\flwss}{0.90}
\begin{figure*}[t]
\centering
    \includegraphics[width=\flwss\linewidth]{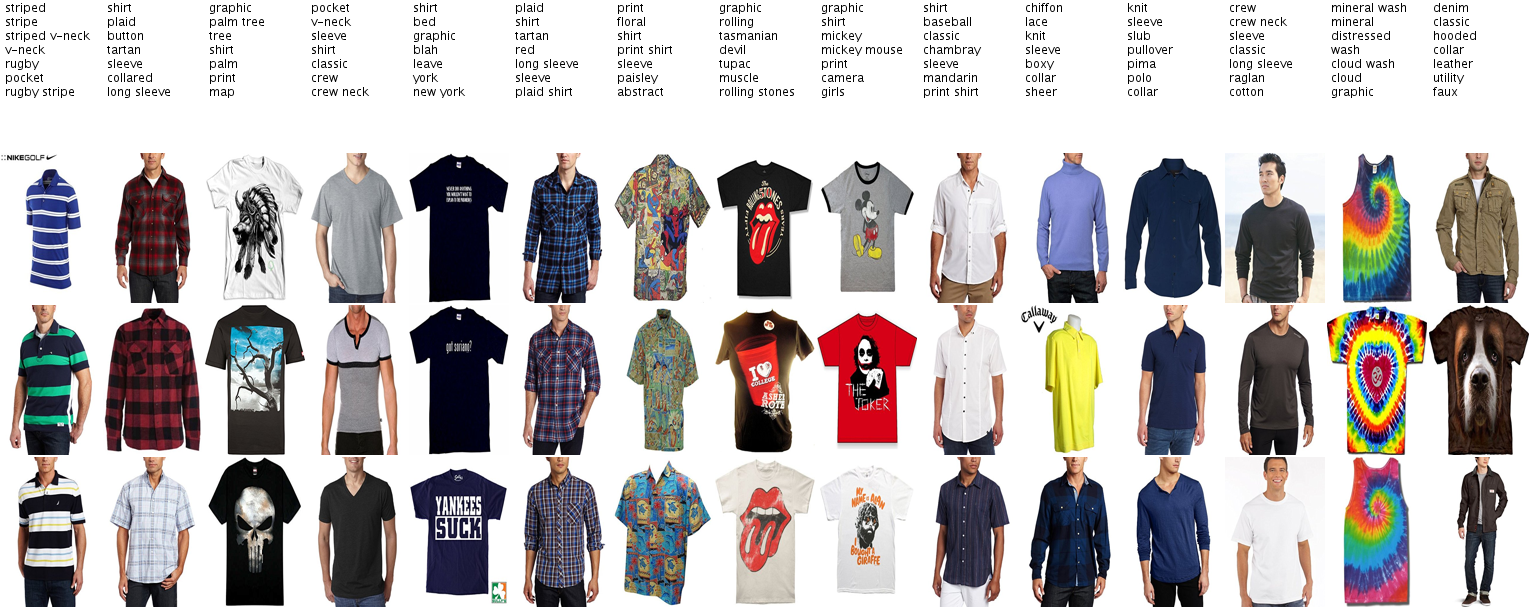}
\caption{The discovered visual styles on the Shirts dataset with their visual signature on top defined by semantic attributes. For discovered styles in Dresses and Tops\&Tees see Figure 3 in the main submission. }
\label{fig:styles_topimgs_supp}
\end{figure*} \begin{figure*}[t]
\centering
\begin{subfigure}[t]{\linewidth}
\centering
    \includegraphics[width=\linewidth]{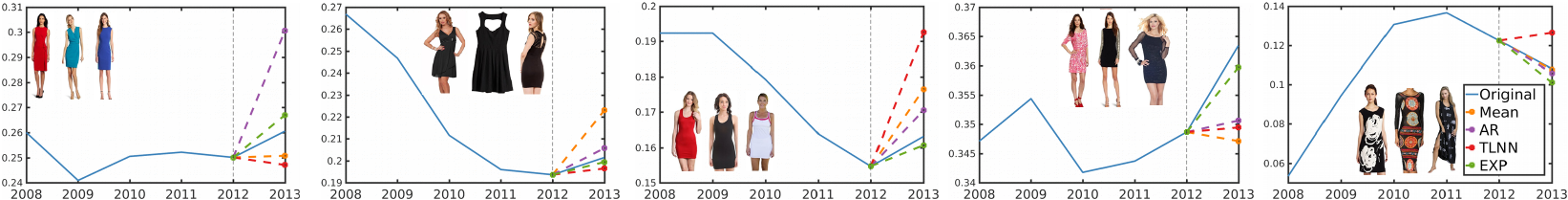}
\caption{Dresses}
\end{subfigure}
\\ \vspace{1.5mm}
\begin{subfigure}[t]{\linewidth}
\centering
    \includegraphics[width=\linewidth]{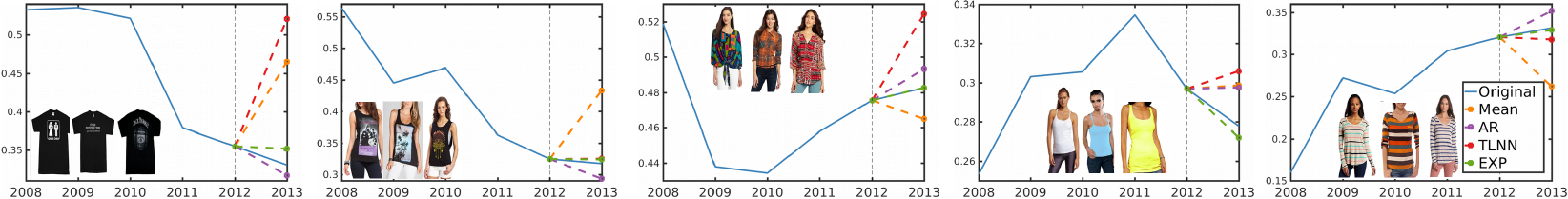}
\caption{Tops\&Tees}
\end{subfigure}
\\ \vspace{1.5mm}
\begin{subfigure}[t]{\linewidth}
\centering
    \includegraphics[width=\linewidth]{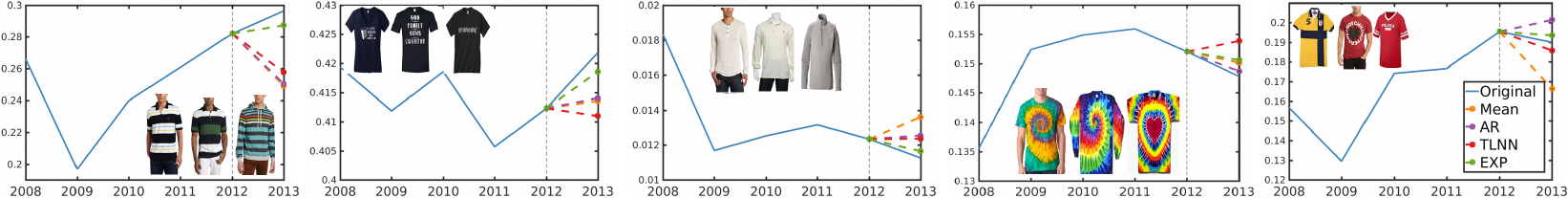}
\caption{Shirts}
\end{subfigure}

\caption{The forecasted popularity of the visual styles in (a) Dresses, (b) Tops\&Tees and (c) Shirts.
Our model (EXP) successfully captures the popularity of the styles in year 2013 with minor errors in comparison to the baselines.}
\label{fig:forecast_supp}
\end{figure*} 
 
\figref{fig:forecast_supp} shows the style popularity forecasts estimated by baselines from the three previous groups in comparison to our approach.
The Naive and NN based forecast models seem to produce larger prediction errors.
Our model performs the best followed by the Autoregressor (AR).
For quantitative comparisons and more detailed discussion see Section 4.2 in the main submission.

\end{document}